\documentclass[runningheads]{llncs}

\usepackage{graphicx}
\usepackage{amsmath}
\usepackage{amsfonts}
\usepackage{hyperref}
\usepackage{subfig}
\usepackage[noend]{algorithm2e}

\begin{document}

\title{Enumerating the k-fold configurations in multi-class classification problems}
\author{Attila Fazekas\inst{1}\orcidID{0000-0001-6893-3067} \and
Gy\"orgy Kov\'acs\inst{2}\orcidID{0000-0003-1736-0988}}
\authorrunning{A. Fazekas and G. Kov\'acs}
\institute{Faculty of Informatics, University of Debrecen, Kassai út 26, Debrecen, H-4028, Hungary
\email{attila.fazekas@inf.unideb.hu}\\
\and
Analitycal Minds Ltd., \'Arp\'ad \'ut 5, Beregsurány, H-4933, Hungary\\
\email{gyuriofkovacs@gmail.com}}

\maketitle         

\begin{abstract}
K-fold cross-validation is a widely used tool for assessing classifier performance. The reproducibility crisis \cite{crisis0} faced by artificial intelligence partly results from the irreproducibility of reported k-fold cross-validation-based performance scores. Recently, we introduced numerical techniques \cite{10} to test the consistency of claimed performance scores and experimental setups. In a crucial use case, the method relies on the combinatorial enumeration of all k-fold configurations, for which we proposed an algorithm in the binary classification case. In this paper, the algorithm is generalized to multi-class classification.

\end{abstract}

\section{Introduction}

Multi-class classification \cite{bishop} is one of the most fundamental tasks in machine learning, with applications spanning various fields of science \cite{hep2} \cite{skin}. The performance of different classification techniques is typically assessed using cross-validation \cite{cv}. Most commonly, k-fold cross-validation \cite{kfold} divides the available data into $k$ disjoint subsets (folds) of equal size (differing at most by 1 element). It uses iteratively every single fold for evaluation and the remaining $k-1$ folds for training the classifier. Evaluation involves calculating various performance scores \cite{2}, such as \emph{accuracy}, defined as the fraction of correctly classified records in the test set. Finally, the performance scores are averaged across all folds to obtain an overall metric of classifier performance \cite{kfold} \cite{cv}.

Recently, several authors have emphasized \cite{crisis0} \cite{crisis1} \cite{cosmetics} that, in many cases, reported performance scores are neither reproducible nor provide a reliable basis for ranking classification techniques in the literature \cite{ehg} \cite{vessel}. The main causes include flawed experimental setups \cite{crisis1}, 'cosmetic adjustments' \cite{cosmetics}, or typographical errors. To address the so-called \emph{reproducibility crisis}, in a previous paper of ours \cite{10}, we developed precise numerical techniques to test whether reported performance scores could result from the claimed experimental setup. In cases where performance scores are averaged over folds in k-fold cross-validation, the proposed techniques require enumerating all possible k-fold configurations regarding the distribution of class labels across folds. For the special case of binary classification, based on integer partitioning \cite{intpart1}, we introduced a recursive algorithm for the enumeration of fold configurations \cite{10}, and successfully applied it to validate reported results in real-world problems \cite{10}.

In this paper, we generalize the method proposed in \cite{10} to enumerate all fold configurations in multi-class classification problems. Beyond its application for consistency testing (analogously to the binary case \cite{10}), the algorithm also enables the analysis of the k-fold configuration space, potentially guiding the selection of the number of folds, especially in small and imbalanced problems.

The paper is organized as follows: Section \ref{sec:problem} provides a precise formulation of the problem, Section \ref{sec:proposed} presents the proposed algorithm, Section \ref{sec:results} discusses some experimental results, and finally, conclusions are drawn in Section \ref{sec:summary}.

\section{Problem formulation}
\label{sec:problem}

Consider a general multi-class classification problem with the training data $(\mathbf{x}_i, y_i)_{i=1}^{N}$, where $\mathbf{x}_i\in\mathbb{R}^{d}$ and $y_i\in\lbrace 0, \dots, m-1\rbrace$ denote the \emph{feature vector}s and corresponding \emph{class label}s, respectively, and $m$ stands for the number of classes. The distribution of class labels is represented by a vector $\mathbf{c}\in\mathbb{Z}_+^{m}$, $\mathbf{c}_i$ being the number of records with class $i$. In a k-fold cross-validation scenario, without stratification, the available data is randomly divided into $k$ disjoint subsets (folds) of approximately equal sizes. Introducing $k_l = (N \mod k)$, there will be $k_l$ folds of size $N_l = \lfloor N/k\rfloor + 1$, and $k_s = k - k_l$ folds of size $N_s = N_l - 1$.
In a particular configuration, let the vector $f_i \in\mathbb{N}^{N_c}$ represent the class distribution in the $i$th fold, $f_i$ sums to either $N_l$ or $N_s$. We introduce the notion of a \emph{fold configuration} for the class distributions of all folds as the matrix $F = [f_0, \dots, f_{k-1}]\in\mathbb{N}^{k\times N_c}$, where $F_{i,j}$ indicates the number of records belonging to class $j$ in fold $i$, naturally, $\sum\limits_{i=0}^{k-1}\sum\limits_{j=0}^{m-1} F_{i,j} = N$. From the application point of view \cite{7}, fold configurations containing the same class distributions in a different order are considered equivalent. Therefore, we introduce the concept of a \emph{standardized fold configuration}, $\hat{F} = [f^{s_0}, \dots, f^{s_k}]$, where the sorting $s_1, \dots, s_k$ ranks the folds primarily by their cardinalities and secondarily by lexicographic ordering. 

For example, consider a classification problem with 3 classes of sizes $c_0 = 10, c_1 = 20, c_2 = 271$. Suppose a 3-fold cross-validation is to be carried out. Then, there will be $k_s=2$ folds with the cardinality $N_s=100$, and $k_l=1$ fold with the cardinality $N_l=101$. One particular fold configuration might consist of $f_0 = (3, 6, 91)$, $f_1 = (5, 7, 89)$, $f_2 = (2, 7, 91)$. The standardized representation of the configuration becomes: $\hat{F} = [[2, 7, 91], [3, 6, 91], [5, 7, 89]]$.

With the definitions introduced above, our goal is to \emph{develop an algorithm for the enumeration of all distinct standardized fold configurations}, given the overall class distribution $\mathbf{c}$ and the number of folds to be used $k$, or equivalently, the sizes of the folds $\mathbf{n}\in\mathbb{Z}_+^{k}$.

\section{The proposed method}
\label{sec:proposed}



Using the notations introduced in Section \ref{sec:problem}, in the special case of 2 classes, and $k$ folds of equal sizes ($N \mod k = 0$), the unique distribution of one class across the folds determines the fold configuration uniquely (the other class needs to complement the folds). Consequently, the problem reduces to the partitioning of the class cardinality $\mathbf{c}_0$ to $k$ parts, each less than or equal to $\lceil N/k\rceil$ -- which is essentially a constrained form of the problem of \emph{integer partitioning} \cite{intpart1} in \emph{enumerative combinatorics} \cite{intpart0}. The enumeration of all integer partitions is extensively researched in the literature, with well-known recursive solutions \cite{intpart0} \cite{intpart1}. (We note that, according to our best knowledge, there are no closed form formulas for the number of distinct partitions.) Since the enumeration of fold configurations in the multi-class case is a generalization of the binary case (and can be considered as a constrained, joint partitioning of integers), we seek for recursive solutions, inspired by the characteristics of existing integer partitioning algorithms.


For clarity and conciseness, in formulating the algorithm, we leverage the generator notation. Multiple modern programming languages (such as Python, JavaScript, C\#, Scala) provide the functionality \emph{yielding from a generator}, which essentially means the generation of one single element of a sequence to be generated, usually denoted by the \emph{yield} keyword. The use of generators with the \emph{yield} operation enables a memory efficient formulation of algorithms and also simplifies implementations. In the pseudo-code, vectors and matrices are indexed from 0 and slicing is denoted with the usual notation (for example, $\mathbf{c}_{1:}$ denoting a vector consisting of all elements of $\mathbf{c}$ except the first).

The proposed algorithm is provided as a pseudo-code in Algorithm~\ref{alg0}, and a Python implementation is available in the repository \url{https://github.com/FalseNegativeLab/mlscorecheck}. 

The function \texttt{Partition22} implements the exactly solvable case of 2 classes and 2 folds (of arbitrary sizes). Having 2 folds of sizes $\mathbf{n}_0$ and $\mathbf{n}_1$ and 2 classes of cardinalities $\mathbf{c}_0$ and $\mathbf{c}_1$ ($\mathbf{n}_0 + \mathbf{n}_1 = \mathbf{c}_0 + \mathbf{c}_1$), the 2-partitions of class $\mathbf{c}_0$ determine the entire partitioning: assuming $F_{0,0} = d$, one can readily see that $F_{0,1} = \mathbf{n}_0 - d$, $F_{1, 0} = \mathbf{c}_0 - d$ and $F_{1, 1} = \mathbf{n}_1 - \mathbf{c}_0 + d$. If $\mathbf{n}_0\neq \mathbf{n}_1$, all configurations are unique. However, if $\mathbf{n}_0 = \mathbf{n}_1$, the configurations get duplicated, hence, in this case the iteration stops when $F_{0,0}$ reaches $\mathbf{c}_0/2$.

The first level of recursive decomposition (\texttt{Partition2M}) is  based on the idea that with 2 folds and $m$ classes of cardinalities $\mathbf{c}_i, i=0, \dots, m-1$, the problem can be treated as a 2-class problem with $\mathbf{c}_0' = \mathbf{c}_0$ and $\mathbf{c}_1' = sum(\mathbf{c}_{1:})$. Generating a configuration $F'$ by \texttt{Partition22} to distribute the first class and the union of the other classes ($\mathbf{c}_1'$) in the two folds and then recursively solving the reduced problem of $\mathbf{n}_0' = \mathbf{n}_0 - F_{0,0}$, $\mathbf{n}_1' = \mathbf{n}_1 - F_{1,0}$, $\mathbf{c}_0' = \mathbf{c}_1$, $\mathbf{c}_1' = sum(\mathbf{c}_{2:}$ for the remaining classes leads to a fold configuration for 2 folds and $m$ classes.

The second level of recursive decomposition (\texttt{PartitionKM}) follows a similar concept but uses the 2-fold m-class solution (\texttt{Partition2M}) for solving the k-fold m-class case. During the generation, we must ensure that k-fold configurations that are identical are skipped, since the order of the folds is irrelevant. In the case of 2 folds, this is ensured by (\texttt{Partition22}), in the case of more than 2 folds, we discard those that are already included by relying on lexicographical sorting (see the condition $\hat{F}_{0,:} > \hat{F}_{1,:})$.

\begin{algorithm}[H]
  \caption{The pseudo-code of enumerating all standardized fold configurations. In the output array $\hat{F}$, $\hat{F}_{i,j}$ represents the cardinality of class $j$ in fold $i$.
  \label{alg0}
  }
  \SetKwProg{Fn}{Function}{:}{}
  \SetKwProg{Gen}{Generator}{:}{}
  \SetKwFunction{PartitionTT}{Partition22}
  \SetKwFunction{PartitionTM}{Partition2M}
  \SetKwFunction{PartitionKM}{PartitionKM}
  \SetKwFunction{Partition}{Partition}
  \SetKw{Yield}{yield}
  \SetKw{Continue}{continue}
  \SetKw{Args}{Args:}
  \SetKw{Yields}{Yields:}
  \SetKw{Description}{Description:}
  \SetKwComment{Comment}{//}{}

  \Gen{\PartitionTT{$n_0$, $n_1$, $c_0$, $\hat{F}$}}{
    \Description{Generate standard configurations for 2 folds with sizes $n_0$ and $n_1$, and 2 classes of cardinalities $c_0$ and $n_0 + n_1 - c_0$ into $\hat{F}$}\\
    \eIf{$n_0 = n_1$}{
      $u \gets min(\lfloor c_0/2\rfloor, n_0)$\;
    }{
      $u \gets min(c_0, n_0)$\;
    }
    
    $l \gets max(c_0 - n_1, 0)$\;
    \For{$i = l \dots u$}{
      $\hat{F}_{0,:} \gets (i, n_0 - i)$\;
      $\hat{F}_{1,:} \gets (c_0 - i, n_1 - c_0 + i)$\;
      \Yield{}
    }
  }
  \Gen{\PartitionTM{$n_0$, $n_1$, $\mathbf{c}$, $\hat{F}$}}{
    \Description{Generate standard configurations for 2 folds of sizes $n_0$ and $n_1$ and $m$ classes of cardinalities $\mathbf{c}\in\mathbb{Z}_{+}^m$} into $\hat{F}$\\
    \While{\PartitionTT{$n_0$, $n_1$, $c_0$, $\hat{F}$} yields}{
      \eIf{$len(\mathbf{c})$ = 2}{
        \Yield{}
      }
      {
        \While{\PartitionTM{$\hat{F}_{0, 1}$, $\hat{F}_{1, 1}$, $\mathbf{c}_{1:}$, $\hat{F}_{1, 1:}$} yields}{
          \Yield{}
        }
      }
    }
  }
  \Gen{\PartitionKM{$\mathbf{n}$, $\mathbf{c}$, $\hat{F}$}}{
    \Description{Generate standard configurations for $k$ folds of sizes $\mathbf{n}\in\mathbb{Z}_{+}^{k}$ and $m$ classes of cardinalities $\mathbf{c}\in\mathbb{Z}_{+}^m$} into the output matrix $\hat{F}\in\mathbb{Z}^{k\times m}$\\
    \While{\PartitionTM{$\mathbf{n}_0$, $sum(\mathbf{n}_{1:})$, $\mathbf{c}$, $\hat{F}$} yields}{
      \eIf{$len(\mathbf{n})$ = 2}{
        \Yield{}
      }
      {
        \While{\PartitionKM{$n_{1:}$, $\hat{F}_{1,:}$, $\hat{F}_{1:,:}$} yields}{
          \If{$\mathbf{n}_0 = \mathbf{n}_1 \wedge \hat{F}_{0, :} > \hat{F}_{1, :}$}{
            \Comment{Avoiding repetitions by enforcing lexicographic ordering}
            \Continue{}
          }
          \Yield{}
        }
      }
    }
  }

\end{algorithm}



\section{Experimental results}
\label{sec:results}


The algorithm has two intended use cases, both related to small multi-class classification problems (which are particularly common in medicine, for example \cite{ehg} \cite{skin}), and in this section we present some preliminary results confirming the applicability of the method. 

The enumeration of all fold configurations enables testing the consistency of reported performance scores when k-fold cross-validation is used for evaluation without stratification \cite{7}. For example, the \emph{post-operative} dataset from the KEEL repository \cite{keel} contains 90 records of 3 classes. With 5-fold cross-validation, the overall number of standardized fold configurations by the proposed method is 2846. Similarly to the approach introduced in \cite{7}, testing the consistency of reported performance scores involves the solution of 2846 small linear integer programming problems, which is feasible in terms of computational demands. Naturally, this use case becomes less tractable as the size of the problems increases; however, further research related to generating the most likely configurations can extend the applicability of the method by statistical approaches.

\begin{figure}[t]
\begin{center}
\subfloat[\label{fig1a}]{
    \includegraphics[width=0.49\textwidth]{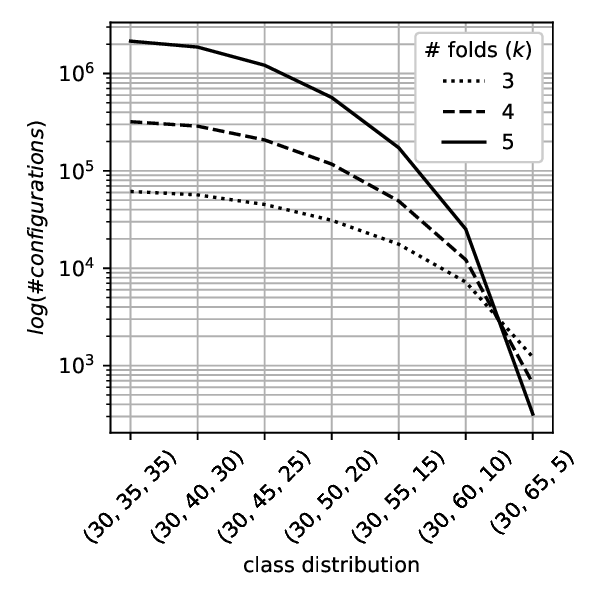}
}
\subfloat[\label{fig1b}]{
    \includegraphics[width=0.49\textwidth]{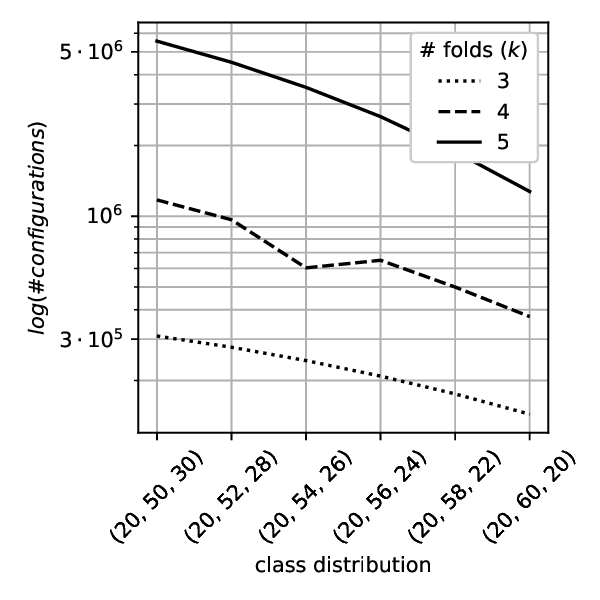}
}
\end{center}
\caption{Experimental results on the number of standardized fold configurations for a problem with 100  records and various class label distributions. As observed in subfigure (a), there is a steep decrease in the number of configurations as the cardinality of the smallest class is reached by the number of folds. Another interesting property of the number of configurations is that irregularities might appear based on the divisibility of class cardinalities by the number of folds - as observable at the configuration (20, 54, 26) with 4 folds in subfigure (b). \label{fig1}}
\end{figure}

The other application is related to the investigation of the relationship between the number of folds ($k$) and the potential variety of fold configurations in certain problems. While there are several rules of thumb, there is no universally accepted method for selecting the number of folds; researchers typically use values in the range of 5 to 10 folds \cite{statlearn}. The variance of cross-validation is usually reduced by repeating the k-fold cross-validation; however, the selection of $k$ influences the variability of the overall population of fold configurations to choose from. The quantitative characterization of the effect of $k$ could provide guidance for its selection, especially in the case of small and imbalanced datasets where the effect of $k$ can be expected to be more significant. We note that the current enumeration does not account for the variability introduced by the permutations of records belonging to the same class; the estimation of which could be a further direction of research. The impact of $k$ on the number of standardized fold configurations of some small datasets is illustrated in Figure \ref{fig1}, and one can observe some anticipated, non-trivial patterns worth further investigation.






\section{Summary and Conclusions}
\label{sec:summary}

In this paper, we proposed a recursive algorithm for the enumeration of fold configurations in multi-class classification problems. The algorithm is a generalization of the one proposed for binary classification in \cite{7} and can also be treated as a constrained generalization and application of the integer partitioning problem of number theory and enumerative combinatorics \cite{intpart0,intpart1}.

The algorithm is directly applicable to the consistency testing of reported performance scores in relatively small multi-class classification problems, following a similar approach to \cite{7}. Additionally, it contributes to the analysis and quantitative characterization of the configuration space of multi-class foldings.

Naturally, the number of fold configurations grows enormously with the complexity of the problem, considering factors such as the number of samples in classes and the number of folds. As a result, this exact algorithm is suitable for relatively small problems only, which are common in fields like medicine. Future developments focused on generating/sampling those fold configurations which are likely to appear in practice (with the highest number of potential permutations of elements), would enable consistency testing \cite{7} to be applicable to larger datasets. Additionally, investigating asymptotic formulas to approximate the number of fold configurations is another valuable research direction. This approach holds the promise of providing combinatorial validation for the choice of $k$ in the range of $5$ to $10$, suggesting that this is the range where the variability of fold configurations in typical problems combinatorially explodes.




\bibliographystyle{splncs04}
\bibliography{PaperLast}

\end{document}